\documentclass[letterpaper]{IEEEtran}

\usepackage{cite}
\usepackage{amsmath,amssymb,amsfonts}
\usepackage{gensymb}
\usepackage{textcomp}
\usepackage{xcolor}
\usepackage{flushend}
\usepackage{algorithm,algpseudocode}
\usepackage{listings}
\DeclareMathOperator*{\argmax}{argmax} 
\usepackage{mathptmx}
\usepackage{graphicx}
\usepackage{textcomp}
\def\BibTeX{{\rm B\kern-.05em{\sc i\kern-.025em b}\kern-.08em
    T\kern-.1667em\lower.7ex\hbox{E}\kern-.125emX}}
\usepackage{makecell}

\usepackage[english]{babel}
\usepackage{amsthm}
\theoremstyle{definition}

\newtheorem{lemma}{Lemma}

\usepackage{caption}
\usepackage{subcaption}
\usepackage{mwe}

\begin{document} 

\title{
Outlier Detection of Poisson-Distributed Targets Using a Seabed Sensor Network

\thanks{*Corresponding author is Mingyu Kim.\\This work was supported in part by the Office of Naval Research under Grant N00014-20-1-2845.}
}

\author{Mingyu Kim$^{1*}$, Daniel J. Stilwell$^2$, and Jorge Jimenez$^3$ \\ 
$^1$Department of Electrical and Computer Engineering,
Georgia Southern University, Statesboro, GA, USA\\

$^2$Bradley Department of Electrical and Computer Engineering,
Virginia Polytechnic Institute and State University, Blacksburg, VA, USA\\

$^3$ Johns Hopkins University Applied Physics Laboratory, Laurel, MD, USA\\
$^*$mkim486@vt.edu}

\maketitle

\IEEEpeerreviewmaketitle

\begin{abstract}
\begin{abstract}
This paper presents a framework for classifying and detecting spatial commission outliers in maritime environments using seabed acoustic sensor networks and log-Gaussian Cox processes (LGCPs). By modeling target arrivals as a mixture of normal and outlier processes, we estimate the probability that a newly observed event is an outlier. We propose a second-order approximation of this probability that incorporates both the mean and variance of the normal intensity function, providing improved classification accuracy compared to mean-only approaches. We analytically show that our method yields a tighter bound to the true probability using Jensen’s inequality. To enhance detection, we integrate a real-time, near-optimal sensor placement strategy that dynamically adjusts sensor locations based on the evolving outlier intensity. The proposed framework is validated using real ship traffic data near Norfolk, Virginia, where numerical results demonstrate the effectiveness of our approach in improving both classification performance and outlier detection through sensor deployment.
\end{abstract}


\end{abstract}

\section{Introduction}

This study addresses the challenge of detecting atypical events in maritime environments using seabed acoustic sensor networks. Such events may include abnormal vessel activity or rare marine mammal movements. Accurately distinguishing these outliers from routine patterns is essential for maritime surveillance, security operations, and ecological monitoring. To build atypical event occurence models, we assume that acoustic sources follow a Poisson distribution and develop a framework to evaluate how newly observed events deviate from established historical behavior.

Motivated by the work in \cite{liu2021event}, which investigates the detection of commission outliers (i.e., excessively frequent events) in continuous-time Poisson processes over a one-dimensional temporal domain, we extend the concept to spatial domains in one dimension. To model the spatial variability of both normal and atypical event distributions, we represent target arrivals using log-Gaussian Cox processes (LGCPs), where the logarithm of the intensity function follows a Gaussian process. This modeling approach has been widely applied in fields such as wildlife ecology \cite{jullum2020estimating}, public health \cite{shirota2017space}, criminology \cite{diggle2005point}, and defense systems \cite{kim2023toward, kim2023role, kim2025variance, kim2025trajectory}. We estimate these spatial intensity functions using the integrated nested Laplace approximation (INLA) \cite{rue2009approximate, lindgren2015bayesian}, a computationally efficient Bayesian inference technique for latent Gaussian models.

To classify outliers, the authors in \cite{liu2021event} propose a proxy for the probability of commission outliers based solely on the mean of the normal intensity function. While this approximation is computationally efficient and effectively captures the central tendency of event occurrence, it does not account for uncertainty, particularly in regions with high variability. To address this limitation, we propose a new approximation that incorporates both the mean and variance of the normal intensity function, enabling a more accurate estimation of the probability of a commission outlier.

Once potential outliers are identified using this proxy, we focus on detecting them by strategically adjusting the sensor network. This need arises from both the approximation error and the inherent uncertainty of commission outliers, which are unknown in practice. To address this, we build on our prior work in near-optimal sensor placement \cite{kim2023toward, kim2025variance}, originally designed to maximize detection probability. In this study, we adapt that framework specifically for outlier detection. By applying Jensen’s inequality, we derive a lower bound on the probability that all outliers are detected and use this bound as the objective for real-time sensor placement.



To validate our method, we use publicly available ship traffic data near Norfolk, Virginia from automatic identification system (AIS) \cite{marinecadastre.gov}, as a baseline for modeling typical maritime activity. We then simulate synthetic outlier events to test classification accuracy and sensor deployment strategies. Our results demonstrate that the proposed framework effectively distinguishes outlier events from normal patterns and enhances detection through adaptive sensor placement.

\subsection*{Contributions}

The primary contributions of this work are twofold. First, we propose an approximation of the probability of commission outliers that accounts for both the mean and variance of the normal intensity function, resulting in improved classification performance, especially in high-uncertainty environments. Second, we analytically characterize the approximation error and demonstrate that our method provides a tighter upper bound compared to the traditional mean-based approximation of outlier probability. Building on this approximation and its theoretical guarantees, we further develop a real-time, near-optimal outlier detection framework through adaptive sensor placement.


The remainder of this paper is organized as follows. Section~\ref{sec:ProblemFormulation} introduces the probability model for commission outliers and derives a closed-form approximation. Section~\ref{sec:GapAnalysis} presents a theoretical analysis of the approximation error and its bounds. Section~\ref{sec:NumericalResults} provides numerical results to demonstrate the performance of the proposed methods. The Appendix includes the proof of the inequality used in the approximation error analysis.

\section{Formulation} \label{sec:ProblemFormulation}
\subsection{Probability of Commission Outlier Target Arrivals}

In this paper, building upon the framework presented in \cite{liu2021event}, we develop a spatially-varying model for outlier target arrivals. Unlike the temporal formulation in \cite{liu2021event}, our spatial model allows events to occur anywhere within a bounded domain. To capture the spatial variability in target arrivals, we employ a log-Gaussian Cox process (LGCP), where the logarithm of the intensity function is modeled as a Gaussian process.

The intensity function $\lambda(s)$ in a spatially inhomogeneous Poisson process represents the expected number of targets per unit area at location $s$. The integral of $\lambda(s)$ over the entire domain quantifies the expected total number of target arrivals.

We model the overall intensity as the sum of two independent components: normal and outlier target arrivals
\begin{align*}
    \lambda(s) = \lambda_0(s) + \lambda_1(s)
\end{align*}
where $\lambda_0(s)$ and $\lambda_1(s)$ denote the intensity functions of normal and outlier arrivals, respectively. Each component follows a log-Gaussian process
\begin{align}
    \log(\lambda_0(s)) &\sim GP(\mu_0(s), \Sigma_0(s,s'))\label{eq: LGCP lambda0} \\
    \log(\lambda_1(s)) &\sim GP(\mu_1(s), \Sigma_1(s,s'))\label{eq: LGCP lambda1}
\end{align}
where $\mu_0(s)$ and $\Sigma_0(s,s')$ are the mean and covariance functions of the normal process. Following Assumption 3.1 in \cite{liu2021event}, we assume that $\lambda_0(s)$ and $\lambda_1(s)$ are independent.

Using this decomposition and Bayes' rule, the probability that an observed event at location $s$ is an outlier, given an unknown but non-negative $\lambda_1(s)$, is expressed as
\begin{align}
    p(Z=1|s) = \mathbb{E}_{\lambda}\left[\frac{\lambda_1(s)}{\lambda(s)}\right] = 1 - \mathbb{E}_{\lambda}\left[\frac{\lambda_0(s)}{\lambda(s)}\right] \label{eq: original probability of outlier target arrival}
\end{align}
where $Z=1$ denotes an outlier and $Z=0$ an inlier.

Since $\lambda_1(s)$ is unknown and $\lambda_0(s)$ is stochastic, the exact computation of \eqref{eq: original probability of outlier target arrival} is intractable. Instead, we approximate the probability of outlier target arrivals using historical data to estimate $\lambda_0(s)$
\begin{align}
    p(Z=1|s) \approx \mathbb{E}_{\lambda_0}\left[\frac{\lambda_1(s)}{\lambda(s)}\right] = 1 - \mathbb{E}_{\lambda_0}\left[\frac{\lambda_0(s)}{\lambda(s)}\right] \label{eq: probability of outlier target arrival}
\end{align}
This approximation serves as a proxy for ranking new observations by their likelihood of being atypical. Because $\lambda_1(s)$ is unknown across the bounded domain $s \in \Psi$, we estimate it through our proposed outlier classification method to compute this probability approximation.

\subsubsection*{Our Proposed Approximation of the Probability of Outlier Target Arrivals}

Inspired by the approach in \cite{liu2021event}, we first consider a mean-based approximation of \eqref{eq: probability of outlier target arrival}
\begin{align}
    P(Z=1|s) \approx 1 - \frac{\mu_0(s)}{\mu_0(s) + \lambda_1(s)} \label{eq: mean approx probability of outlier target arrival}
\end{align}
where $\mu_0(s) = \mathbb{E}_{\lambda_0}[\lambda_0(s)]$ is the expected normal intensity. In our approach, we derive a second-order Taylor approximation of \eqref{eq: probability of outlier target arrival} around the mean of the normal intensity function $\mu_0(s)$
\begin{align}
    P(Z=1|s) \approx 1 - \frac{\mu_0(s)}{\mu_0(s) + \lambda_1(s)} + \frac{\lambda_1(s) \sigma_0(s)^2}{(\mu_0(s) + \lambda_1(s))^3} \label{eq: our approx probability of outlier target arrival}
\end{align}
where $\sigma_0(s)$ is the standard deviation of $\lambda_0(s)$. This expression captures the effect of variability in the normal arrival rate and improves the approximation of the outlier probability. The improvement is described in detailed in Sec.~\ref{sec:GapAnalysis}. We estimate $\mu_0(s)$ and $\sigma_0(s)$ from historical normal target arrival data using the integrated nested Laplace approximation (INLA) method.

\begin{algorithm}[b!]
  \caption{Classification of Potential Commission Outliers}
  \label{algo:outlierClassification}
  \begin{algorithmic}[1]
    \State \textbf{Input:} 
    \State \hspace{1em} $\mu_N$: estimated mean number of normal target arrivals
    \State \hspace{1em} $\sigma_N$: estimated standard deviation of normal target arrivals
    \State \hspace{1em} $W = \{\omega_1, \dots, \omega_K\}$: observed target arrival locations
    \State \textbf{Output:} 
    \State \hspace{1em} $\hat{W} = \{\hat{\omega}_1, \dots, \hat{\omega}_L\}$: classified potential commission outliers ($L \leq K$)

    \State \textbf{Initialization:}
    \State \hspace{1em} $\hat{W} \gets \{\}$ \Comment{Initialize empty set for classified outliers}
    
    \For{$j = 1$ to $K$} \Comment{Iterate over observed arrivals}
      \State Sample $u_1 \sim \mathcal{U}(0, 1)$

      \If{$u_1 < P(X < K)$} \Comment{Compare with CDF of $X \sim \mathcal{N}(\mu_N, \sigma_N^2)$}
        \State Sample $u_2 \sim \mathcal{U}(0, 1)$

        \If{$u_2 < P(Z = 1 \mid s = \omega_j)$} 
        \Comment{Evaluate outlier probability at location $\omega_j$ using \eqref{eq: mean approx probability of outlier target arrival} or \eqref{eq: our approx probability of outlier target arrival}}
          \State $\hat{W} \gets \hat{W} \cup \{\omega_j\}$ 
          \Comment{Classify $\omega_j$ as a potential outlier}
        \EndIf
      \EndIf
    \EndFor
  \end{algorithmic}
\end{algorithm}

\subsection{Algorithm for Potential Commission Outlier Classification}

To evaluate \eqref{eq: our approx probability of outlier target arrival}, we need an estimate of $\lambda_1(s)$. Since $\lambda_1(s)$ corresponds to atypical (outlier) target arrivals, we propose a dynamic classification method to identify potential commission outliers based on observed data.

As outlined in Algorithm~\ref{algo:outlierClassification}, the classification process is guided by the estimated mean and standard deviation of the number of normal target arrivals per unit time $T_c$, denoted by $\mu_N$ and $\sigma_N$. These are computed by generating a large number of samples (e.g., $>10{,}000$) from the posterior of the normal intensity function
\[
\log(\hat{\lambda}_{0,q}(s)) \sim GP(\mu_0(s), \Sigma_0(s, s'))
\]
where $q=1,2,...$ is an index of an sampled intensity function, and integrating each sampled intensity function over the spatial domain $\Psi$
\[
\hat{\Lambda}_{0,q} = \int_{\Psi} \hat{\lambda}_{0,q}(s) \, ds
\]
which represents the number of ship arrivals by a sampled intensity function.
Assuming the distribution of the number of normal arrivals is approximately Gaussian, we estimate $\mu_N$ and $\sigma_N$ from the empirical distribution of $\hat{\Lambda}_0$ based on a large number of samples.

To determine whether a given set of $K$ observed target arrivals includes any commission outliers, we use the cumulative distribution function (CDF) of a normal distribution parameterized by $\mu_N$ and $\sigma_N$, as implemented in Line~11 of Algorithm~\ref{algo:outlierClassification}. As the number of observed arrivals increases, the probability that the set contains at least one commission outlier also tends to increase, reflecting the nature of rare but excessive occurrences in such processes. This test involves a random threshold $u_1$ to ensure probabilistic decision-making. If outliers are detected, each individual arrival is further evaluated using the approximations of the outlier probability in \eqref{eq: mean approx probability of outlier target arrival} and \eqref{eq: our approx probability of outlier target arrival}. Based on this evaluation, each target arrival is classified as either a potential commission outlier or not, as described in Lines 12–13 of the algorithm.

In this work, we assume that all target arrivals are observable, such as those visible from a lighthouse. However, in practical settings, only sensors deployed in close proximity to the targets can evaluate whether an arrival is likely to be an outlier. This setup enables each sensor to contribute to real-time classification, progressively refining the estimate of the outlier intensity function $\lambda_1(s)$. Consequently, real-time operation of sensor placement for outlier classification and detection becomes essential, not only to reduce the impact of misclassification or approximation error, but also to improve the system’s ability to detect future anomalies more accurately.

\subsection{Real-Time and Near-Optimal Sensor Placement for Commission Outlier Target Arrival Detection}

Since commission outliers are inherently unknown and approximation errors exist in estimating the probability of outlier target arrivals, some inliers may be misclassified as outliers, and true outliers may go undetected. To mitigate this issue, we estimate the outlier intensity function using INLA based on the set of classified potential outliers and apply a real-time, near-optimal sensor placement strategy. This allows the sensor network to continuously adjust its configuration and improve detection performance over time.

\subsubsection{Void Probability of Potential Commission Outlier Target Arrivals}

Once potential commission outlier target arrivals are classified and $\lambda_1(s)$ is estimated, we aim to strategically place sensors to minimize the probability of missing these atypical events. To achieve this, we adopt our real-time target detection framework introduced in \cite{kim2023role, kim2023toward, kim2025variance} for 1-D domains and \cite{kim2025trajectory} for 2-D domains.

For the 1-D case studied in \cite{kim2023toward}, the void probability (i.e., the probability that sensors do not miss any commission outlier targets) is expressed as
\begin{align}
    \mathbb{E}_{\lambda_1}\left[e^{-\Lambda_1(\mathbf{a},\Psi)} \right] \label{eq: original objective fn}
\end{align}
where $\Lambda_1(\mathbf{a}, \Psi)$ represents the expected number of undetected outlier targets given a sensor network $\mathbf{a} = \{a_1, a_2, \dots, a_M\}$ with each sensor located at $a_i \in \Psi$. This is defined by:
\begin{align}
    \Lambda_1(\mathbf{a}, \Psi) = \frac{T}{T_c} \int_{\Psi} \lambda_1(s) \prod_{i=1}^{M} (1 - \gamma(s, a_i)) \, ds \label{eq: num undetected targets}
\end{align}
Here, $T$ and $T_c$ denote the time period of interest and the duration of the historical data collected for estimation, representing unit time, respectively. 
For simplicity, we denote $\Lambda_1(\mathbf{a}, \Psi)$ as $\Lambda_1$ for the remainder of this paper.

The term $\gamma(s, a_i)$ represents the probability of detecting a target at location $s$ by a sensor located at $a_i$. Consequently, the joint probability that all $M$ sensors fail to detect a target at $s$ is
\begin{align}
    \prod_{i=1}^{M} (1 - \gamma(s, a_i)) \label{eq:prob. failing detection}
\end{align} 

\subsubsection{Searching for Near-Optimal Sensor Locations in Real-Time}

Due to the stochastic nature of \eqref{eq: original objective fn} as defined in \eqref{eq: LGCP lambda1}, direct evaluation is computationally intensive. To overcome this, we adopt the approximation from \cite{kim2025variance}
\begin{align}
    \mathbb{E}_{\lambda_1}[e^{-\Lambda_1}] \approx e^{-\mathbb{E}_{\lambda_1}[\Lambda_1]} \left(1 + \frac{1}{2} \sigma_{\Lambda_1}^2 \right) \label{eq: approx solution}
\end{align}
where the variance term is given by:
\[
    \sigma_{\Lambda_1}^2 = \mathbb{E}_{\lambda_1}[\Lambda_1^2] - \left( \mathbb{E}_{\lambda_1}[\Lambda_1] \right)^2
\]
To evaluate this approximation, we separately compute the first and second moments of $\Lambda_1$. Using this expression, we search for near-optimal sensor placements by solving
\begin{align}
    \mathbf{a}^\star = \argmax_{\mathbf{a} \subset \mathbf{A}} \, e^{-\mathbb{E}_{\lambda_1}[\Lambda_1]} \left(1 + \frac{1}{2} \sigma_{\Lambda_1}^2 \right) \label{eq:optimalSensorLocations}
\end{align}
where $\mathbf{A}$ is the set of all possible sensor locations.

As the number of sensors $M$ increases, the search space for optimal placement grows combinatorially, making brute-force optimization infeasible for real-time applications. To address this, we adopt a greedy selection algorithm as described in \cite{kim2023toward,kim2025variance}. This approach leverages the properties of submodularity, monotonicity, and non-negativity, which allow us to obtain a performance guarantee relative to the true optimum
\[
    \text{Objective Greedy} \geq \left(1 - \frac{1}{e} + \alpha\right) \cdot \text{Objective Optimal}
\]
where $\alpha$ is a positive constant such that $0 < \alpha \leq \frac{1}{e}$, representing the improvement gained by incorporating the variance term in \eqref{eq: approx solution}.

\section{Analysis of Approximation Error in the Probability of Outlier Target Arrivals} \label{sec:GapAnalysis}

In our proposed method, the approximation of the probability of outlier target arrival in \eqref{eq: our approx probability of outlier target arrival} incorporates the uncertainty in the normal intensity function $\lambda_0(s)$. However, it is not immediately evident whether this approximation is always better than the simpler approximation that considers only the mean $\mu_0(s)$, as shown in \eqref{eq: mean approx probability of outlier target arrival}. In this section, we analytically show that our proposed second-order approximation provides a closer estimate to the true probability in \eqref{eq: probability of outlier target arrival} given $\lambda_1(s)$.

We begin by noting that the function
\[
\frac{\lambda_0(s)}{\lambda_0(s)+\lambda_1(s)}
\]
is concave with respect to $\lambda_0(s)$ for a fixed $\lambda_1(s) > 0$. Thus, by Jensen's inequality, we have
\begin{align*}
    \mathbb{E}_{\lambda_0}\left[\frac{\lambda_0(s)}{\lambda_0(s)+\lambda_1(s)}\right] \leq \frac{\mu_0(s)}{\mu_0(s)+\lambda_1(s)}
\end{align*}
This means that the approximation using only the mean of $\lambda_0(s)$ overestimates the expected value. We define the approximation error (or Jensen's gap) as
\begin{align*}
    J(\lambda_0(s), \lambda_1(s)) = \frac{\mu_0(s)}{\mu_0(s) + \lambda_1(s)} - \mathbb{E}_{\lambda_0}\left[\frac{\lambda_0(s)}{\lambda_0(s) + \lambda_1(s)}\right]
\end{align*}
which is always non-negative due to the concavity of the function. For simplicity, we drop the dependence on $s$ in the notation and use $\lambda_0$, $\mu_0$, $\sigma_0$, $\lambda_1$, $J$, and $\widetilde{J}$ throughout the rest of the sections.

By applying Theorem 2 from \cite{liao2018sharpening} for the case of a concave function, the Jensen gap can be bounded as

\begin{align}
    J_{low} \leq J \leq J_{up}  \label{eq: Jensen gap derivation}
\end{align}
where
\begin{align*}
    J_{low} =  \inf_{\lambda_0 \in [0, \infty)} \frac{\lambda_1\sigma_0^2}{(\lambda_0 + \lambda_1)(\mu_0 + \lambda_1)^2}\\
    J_{up} =\sup_{\lambda_0 \in [0, \infty)} \frac{\lambda_1\sigma_0^2}{(\lambda_0 + \lambda_1)(\mu_0 + \lambda_1)^2}
\end{align*}
Because the function $\frac{\lambda_1\sigma_0^2}{(\lambda_0 + \lambda_1)(\mu_0 + \lambda_1)^2} $ is monotonic decreasing with respect to $\lambda_0 \in [0, \infty)$, we get its minimum and maximum as $\lambda_0$ goes to $\infty$ and zero, respectively. Therefore, we can bound the Jensen's gap

\begin{align}
    0 \leq J(\lambda_0, \lambda_1) \leq \frac{\sigma_0^2}{(\mu_0 + \lambda_1)^2} \label{eq: Jensens's inequality 1}
\end{align}

\subsubsection*{Approximation Error $\Tilde{J}$ of Our Proposed Probability of Outlier  }

We define the Jensen’s gap for our proposed second-order approximation of the probability of commission outlier target arrival as follows
\begin{align}
    \widetilde{J}(\lambda_0, \lambda_1) 
    &= 
     \left( \frac{\mu_0}{\mu_0 + \lambda_1} 
    - \frac{\lambda_1 \sigma_0^2}{(\mu_0 + \lambda_1)^3} \right) - \mathbb{E}_{\lambda_0}\left[\frac{\lambda_0}{\lambda_0 + \lambda_1}\right] \nonumber 
\end{align}
Thus, using \eqref{eq: Jensens's inequality 1}, the corresponding bounds for this approximation error are
\begin{align}
    -\frac{\lambda_1 \sigma_0^2}{(\mu_0 + \lambda_1)^3} \leq \widetilde{J}(\lambda_0, \lambda_1) \leq \frac{\sigma_0^2}{(\mu_0 + \lambda_1)^2} - \frac{\lambda_1 \sigma_0^2}{(\mu_0 + \lambda_1)^3}
    \label{eq: Jensens's inequality 2}
\end{align}

\begin{lemma}
Let $\xi$ be a positive random variable with expectation $\mathbb{E}_{\lambda}[\xi] > 0$ and variance $\sigma_{\xi}^2 > 0$, and let $c \geq 0$ be an unknown constant. Then, the following inequality holds:
\begin{align}
    \max\left(|J(\xi,c)|\right) > \max\left(|\widetilde{J}(\xi,c)|\right),
\label{eq: proposed inequality}
\end{align}
where $J(\xi,c)$ and $\widetilde{J}(\xi,c)$ denote the approximation errors in estimating the probability of outlier target arrivals using \eqref{eq: mean approx probability of outlier target arrival} and \eqref{eq: our approx probability of outlier target arrival}, respectively.
\end{lemma}

Lemma~1 establishes that our proposed second-order approximation \eqref{eq: our approx probability of outlier target arrival} yields a smaller approximation error (in magnitude) compared to the mean-only approximation in \eqref{eq: mean approx probability of outlier target arrival}. The proof of Lemma~1 is provided in the Appendix.

\begin{figure}[b!]
\centering
\includegraphics[scale=0.95]{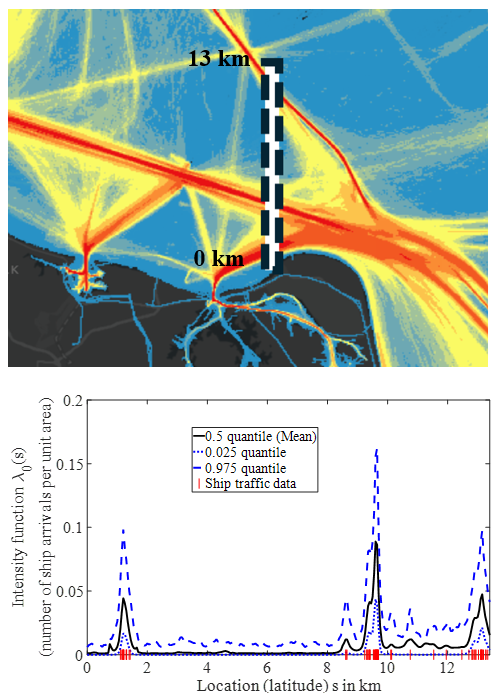}
\caption{Top: Heatmap of ship traffic data near Norfolk, Virginia, USA \cite{marinecadastre.gov}. Bottom: Estimated intensity function along the white line segment in the top image using ship arrival data from April 1st, 2021.}
\label{fig:1}
\end{figure}

\section{Numerical Results}\label{sec:NumericalResults}

In this section, we evaluate the proposed method using real-world ship traffic data collected near Hampton Roads in Norfolk, Virginia, USA. The dataset is publicly available from \cite{marinecadastre.gov}, and a corresponding heatmap of ship traffic is shown in the top panel of Fig.~\ref{fig:1}. In the heatmap, red and yellow represent areas of high and moderate ship traffic, respectively, while blue and black denote ocean and land regions with no recorded ship traffic. With the collected normal ship arrival data, we estimate the intensity function using INLA as shown in the bottom of Fig.~\ref{fig:1}.

We use ship arrival data from 2021 across different time windows (single day, one week, and one month) to demonstrate the effectiveness of our approach in (i) classifying potential commission outliers and (ii) identifying near-optimal sensor placements for detecting them. Specifically, we assess our method's ability to identify commission outlier (i.e., excessive) target arrivals using historical ship traffic data when new data becomes available. To evaluate classification performance, we synthetically generate outlier arrivals and superimpose them on sampled normal target arrival patterns. 

Our analysis focuses on a one-dimensional (1-D) scenario in which ship arrivals are modeled along a 13-km-long line segment (see top of Fig.~\ref{fig:1}). Although this study concentrates on the 1-D case, our proposed framework based on LGCPs is generalizable to higher-dimensional settings, such as the 2-D case explored in \cite{kim2025trajectory}.

\subsection{Estimation of Normal Target Arrivals}

Using historical data, we estimate the mean and standard deviation of the typical number of ship arrivals along the line segment. This is done by sampling a large number $H$ realizations of the normal intensity function, denoted $\hat{\lambda}_{0,q}(s)$, via INLA. The mean number of ship arrivals is computed as
\begin{align*}
    \mu_N = \frac{1}{H} \sum_{q=1}^{H} \frac{1}{T_c} \int_{\Psi} \hat{\lambda}_{0,q}(s) \, ds
\end{align*}
Assuming the number of ship arrivals follows a Gaussian distribution, we compute the standard deviation $\sigma_N$ from these $H$ samples. For instance, using ship traffic data from April 1st, 2021 (shown in the bottom panel of Fig.~\ref{fig:1}), we obtain $\mu_N = 4.03$ and $\sigma_N = 9.99$.

To sample the intensity function, we first estimate the intensity function by implementing INLA. We use the \texttt{inlabru} package in R~\cite{bachl2019inlabru}, which is built upon the R-INLA framework~\cite{lindgren2015bayesian}. The underlying model assumes a Gaussian process prior with zero mean and a Matérn covariance function:
\begin{equation*}
    k(s, s') = \sigma_u^2 \cdot \frac{2^{1 - \zeta}}{\Gamma(\zeta)} \left(\kappa \| s - s' \| \right)^{\zeta} K_{\zeta} \left(\kappa \| s - s' \|\right)
\end{equation*}
where $s$ and $s'$ denote spatial coordinates within the domain, $\sigma_u^2$ is the marginal variance, and $\zeta > 0$ controls smoothness. The scale parameter $\kappa = \sqrt{8\zeta}/\beta$ depends on the spatial range parameter $\beta > 0$, and $K_{\zeta}(\cdot)$ denotes the modified Bessel function of the second kind (see \cite{rpackagedocumentation_2019} for implementation details).

\begin{figure}
\centering
\includegraphics[scale=0.28]{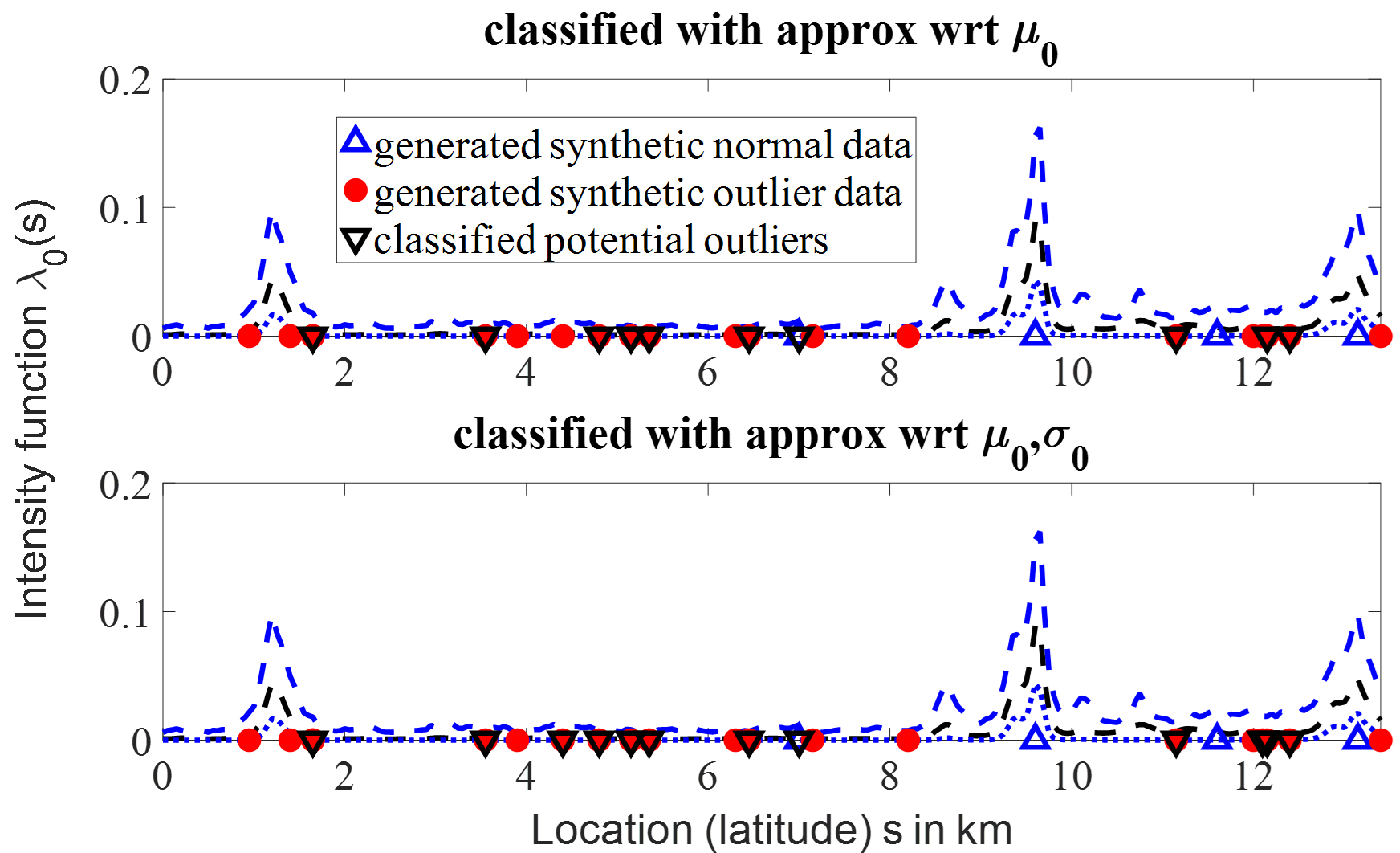}
\caption{Classification of potential commission outliers from a newly collected dataset using Algorithm~\ref{algo:outlierClassification} based on (top) \eqref{eq: mean approx probability of outlier target arrival} and (bottom) \eqref{eq: our approx probability of outlier target arrival}.}
\label{fig:2}
\end{figure}

\subsection{Generation of Synthetic Ship Arrival Data Including Outliers }

To apply Algorithm \ref{algo:outlierClassification}, we need a set of new observed data. In this study, we generate both normal and outlier ship arrivals. Fig.~\ref{fig:2} illustrates an example of synthetic target arrival data that includes both normal and outlier ship arrivals. 

The normal ship arrivals are generated using a sampled intensity function $\hat{\lambda}_{0,q}(s)$, which determines the expected number of arrivals along the spatial domain. To simulate normal ship arrival locations
\begin{enumerate}
    \item We first normalize the sampled intensity function $\hat{\lambda}_{0,q}(s)$ over the domain.
    \item Then, we uniformly draw candidate locations across the domain.
    \item For each candidate location, we generate a random number in the interval $[0, 1]$.
    \item If the random number is less than the normalized intensity value at that location, we accept the location as a ship arrival.
    \item This process is repeated until the total number of accepted arrivals matches the expected count derived from the intensity.
\end{enumerate}
Through this process, one set of normal ship arrivals is generated, as illustrated in Fig.~\ref{fig:2}, where blue triangles indicate their locations around the 10~km and 13~km positions along the line segment.

\begin{table}[t!]
\caption{Success rate of potential commission outlier classification by \eqref{eq: mean approx probability of outlier target arrival} and \eqref{eq: our approx probability of outlier target arrival} when Algorithm \eqref{algo:outlierClassification} is iteratively applied ($\#$ of iteration : 10,000).}
\begin{center}
\begin{tabular}{|c|c|c|}
\hline

\textbf{ \thead{Time period \\ when data collected}} & \textbf{ \thead{Avg success rate \\ by \eqref{eq: mean approx probability of outlier target arrival}}}& \textbf{ \thead{Avg success rate \\ by \eqref{eq: our approx probability of outlier target arrival} }}  \\
\hline

Apr $1^{st}$ & \thead{0.3355} & 0.3758  \\
\hline
Apr $2^{nd}$& \thead{0.3838} & 0.3909  \\
\hline
May $11^{st}$& \thead{0.4200} & 0.4280  \\
\hline
May $12^{th}$ & \thead{0.5190} & 0.5450  \\
\hline
June $21^{st}$ & \thead{0.4506} & 0.4528  \\
\hline
June $22^{nd}$ & \thead{0.3769} & 0.4087  \\
\hline
Apr Week 1 & \thead{0.5249} & 0.5270  \\
\hline
Apr Week 2 & \thead{0.4884} & 0.4894 \\
\hline
May Week 2 & \thead{0.5060} & 0.5070  \\
\hline
May Week 3 & \thead{0.4825} & 0.4846  \\
\hline
June Week 3 & \thead{0.4459} & 0.4463  \\
\hline
June Week 4 & \thead{0.5059} & 0.5063  \\
\hline
April & \thead{0.5571} & 0.5572 \\
\hline
May & \thead{0.4611} & 0.4614 \\
\hline
June & \thead{0.4420} & 0.4422 \\
\hline
July & \thead{0.5256} & 0.5260 \\
\hline
Aug & \thead{0.4368} & 0.4370  \\
\hline
Sep & \thead{0.4886} & 0.4891  \\
\hline

\end{tabular}
\label{tab1}
\end{center}
\end{table}

In our example, we use a uniform intensity function over the bounded domain for generating commission outlier target arrivals, defined as $\lambda_1(s) = 0.00154$. While this intensity is unknown during the actual outlier classification process, it corresponds to an expected count of 20 outlier ship arrivals per unit time ($T = 1$ hour) along the 13~km line segment. As shown in Fig.~\ref{fig:2}, the red circles represent the generated commission ship arrivals, which are uniformly distributed along the line segment.

\subsection{Classification of Potential Commission Outliers}
Assuming full observability of ship arrival data, we treat both normal and outlier arrivals (shown as blue triangles and red circles in Fig.~\ref{fig:2}) as an unclassified observations. To classify them, we compute the probability of an outlier arrival at location $s$ using the estimated parameters $\mu_0(s)$ and $\sigma_0(s)$ (from INLA), along with an initial guess for $\lambda_1(s)$. Specifically, we initialize $\lambda_1(s)$ as $0.00075$, corresponding to 10 outlier arrivals along the line segment per hour. Even though the outlier target arrivals are generated using a uniform distribution in our simulation, we model $\lambda_1(s)$ as a log-Gaussian Cox process (LGCP) to allow flexibility beyond the uniform case. Since our goal is real-time operation, we continuously classify potential outliers and update the estimate of $\lambda_1(s)$ within the LGCP framework as new data become available.

Fig.~\ref{fig:2} compares the classification results using the mean-based approximation in \eqref{eq: mean approx probability of outlier target arrival} (top) and our proposed second-order approximation in \eqref{eq: our approx probability of outlier target arrival} (bottom). The former considers only the mean of the normal intensity $\mu_0(s)$ and may misclassify arrivals in regions where the mean is low but the variance is high. In contrast, our method explicitly incorporates the variance term $\sigma_0(s)^2$, allowing for improved classification under uncertainty.

\begin{figure}
\centering
\includegraphics[scale=0.48]{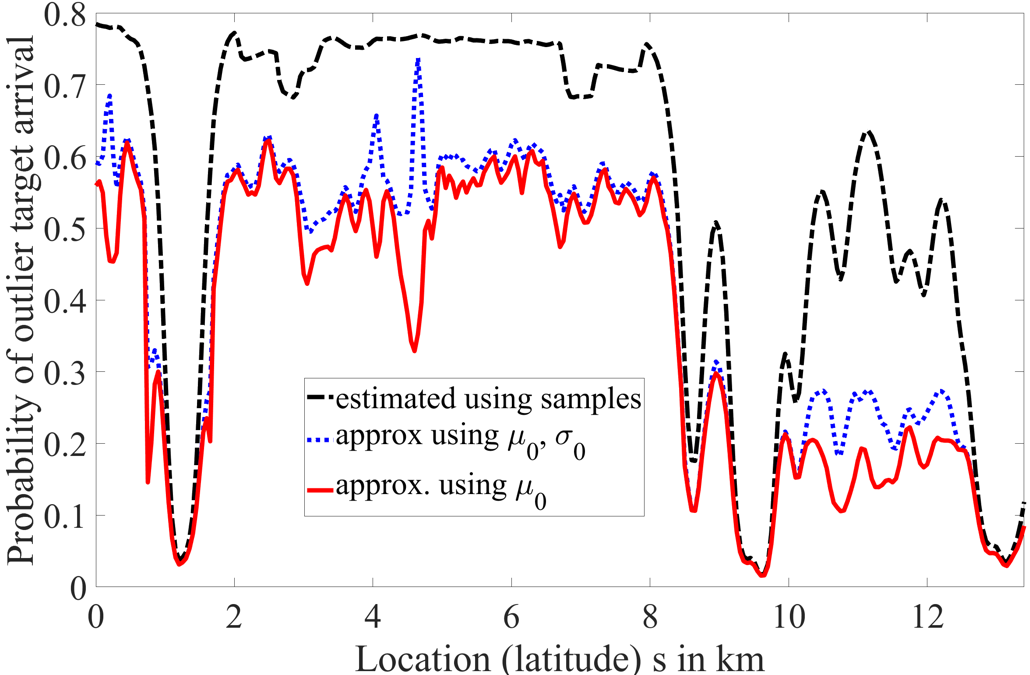}
\caption{Estimated probability of outlier target arrivals for the ship traffic data of April 1st, 2021. The black dotted-dashed curve shows the Monte Carlo estimate using 10,000 samples. The red curve shows the approximation using only the mean of $\lambda_0(s)$ (based on \eqref{eq: mean approx probability of outlier target arrival}), while the blue dotted curve incorporates both the mean and standard deviation of $\lambda_0(s)$ (based on \eqref{eq: our approx probability of outlier target arrival}).}
\label{fig:3}
\end{figure}

For instance, at the outlier near 4.2~km in Fig.~\ref{fig:1}, the mean of the normal intensity function exhibits a small hump that is relatively elevated compared to nearby regions between 2 and 8~km. As a result, the mean-only approximation in \eqref{eq: mean approx probability of outlier target arrival} underestimates the probability of an outlier at this location. In contrast, our proposed approximation in \eqref{eq: our approx probability of outlier target arrival}, which incorporates the variance term, provides a more accurate estimate since it has relatively high variance. Specifically, the outlier probability given by our method is approximately 0.53 higher than that of the mean-only approximation, as shown in Fig.~\ref{fig:3}. This demonstrates the benefit of accounting for uncertainty through the variance term in improving classification performance.


In addition, we evaluate our method using 6 different one-day, one-week, and one-month datasets. Table~\ref{tab1} reports the performance improvements of our proposed approximation over the mean-only method. The improvements are $4.58\%$, $0.37\%$, and $0.06\%$ for the day, week, and month-long datasets, respectively. This indicates that when data is sparse (e.g., shorter collection periods), the variance in the intensity function is more influential, and accounting for this uncertainty tends to yield better approximation performance.

\subsection{Approximation Errors of Outlier Probability (Jensen Gap)}

 \begin{figure}
\centering
\includegraphics[scale=0.48]{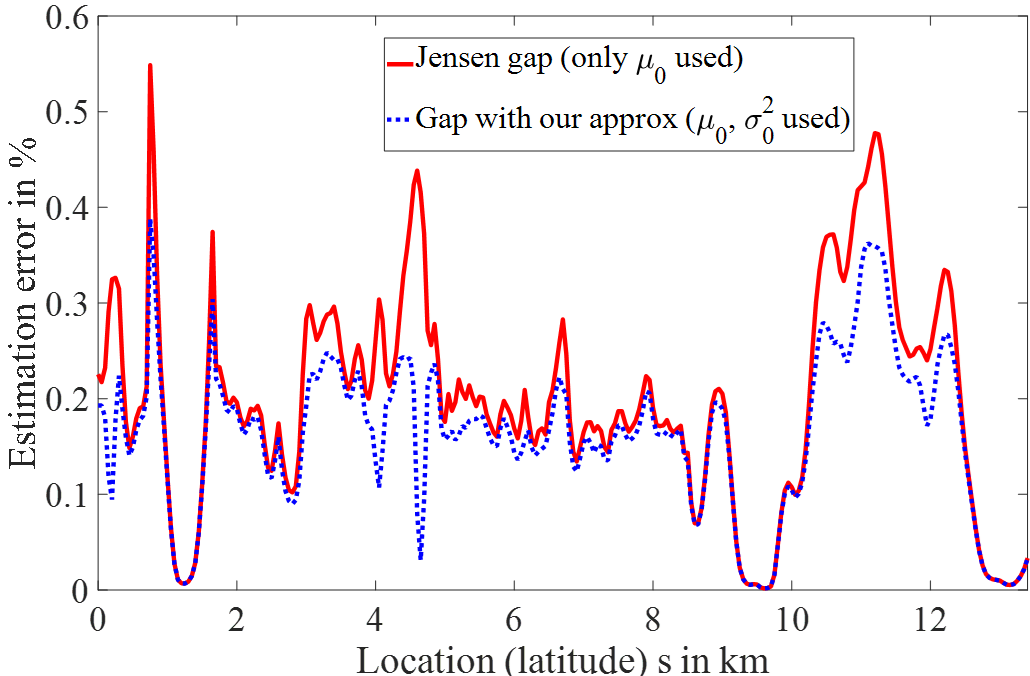}
\caption{Approximation errors (Jensen gaps) in estimating the probability of outlier target arrivals at $s$ corresponding to Fig.~\ref{fig:3}. The figure compares the error of the mean-only approximation and the proposed second-order approximation.}
\label{fig:4}
\end{figure}

As shown in Lemma~1, the approximation error from our proposed method in \eqref{eq: our approx probability of outlier target arrival} is guaranteed to be less than or equal to the error from the mean-only approximation in \eqref{eq: mean approx probability of outlier target arrival}. 

For the example using ship traffic data from April 1st, 2021, Fig.~\ref{fig:3} compares the estimated probabilities of outlier target arrivals. The black dotted-dashed curve represents the Monte Carlo estimate, obtained by sampling $H (>10,000)$ realizations of the normal intensity function and averaging the corresponding probabilities using
\begin{align*}
    P(Z = 1 \mid s) \approx \frac{1}{H} \sum_{q=1}^{H} \left( \frac{\hat{\lambda}_{0,q}(s)}{\hat{\lambda}_{0,q}(s) + \lambda_1(s)} \right)
\end{align*}
The blue dotted and red curves represent approximations of the probability of commission outliers.

Correspondingly, Fig.~\ref{fig:4} illustrates the Jensen gap (i.e., the approximation error) across the spatial domain for both approximation methods, computed as the difference between the Monte Carlo estimate (black curve) and the approximations using the mean-only (red curve) and our proposed (blue curve) methods in Fig.~\ref{fig:3}. Our proposed method achieves a $21.37\%$ reduction in average estimation error over the domain $\Psi$ compared to the mean-only approach.

\subsection{Near-Optimal Sensor Placement for Detection of Commission Outliers}

\begin{figure}
\centering
\includegraphics[scale=0.28]{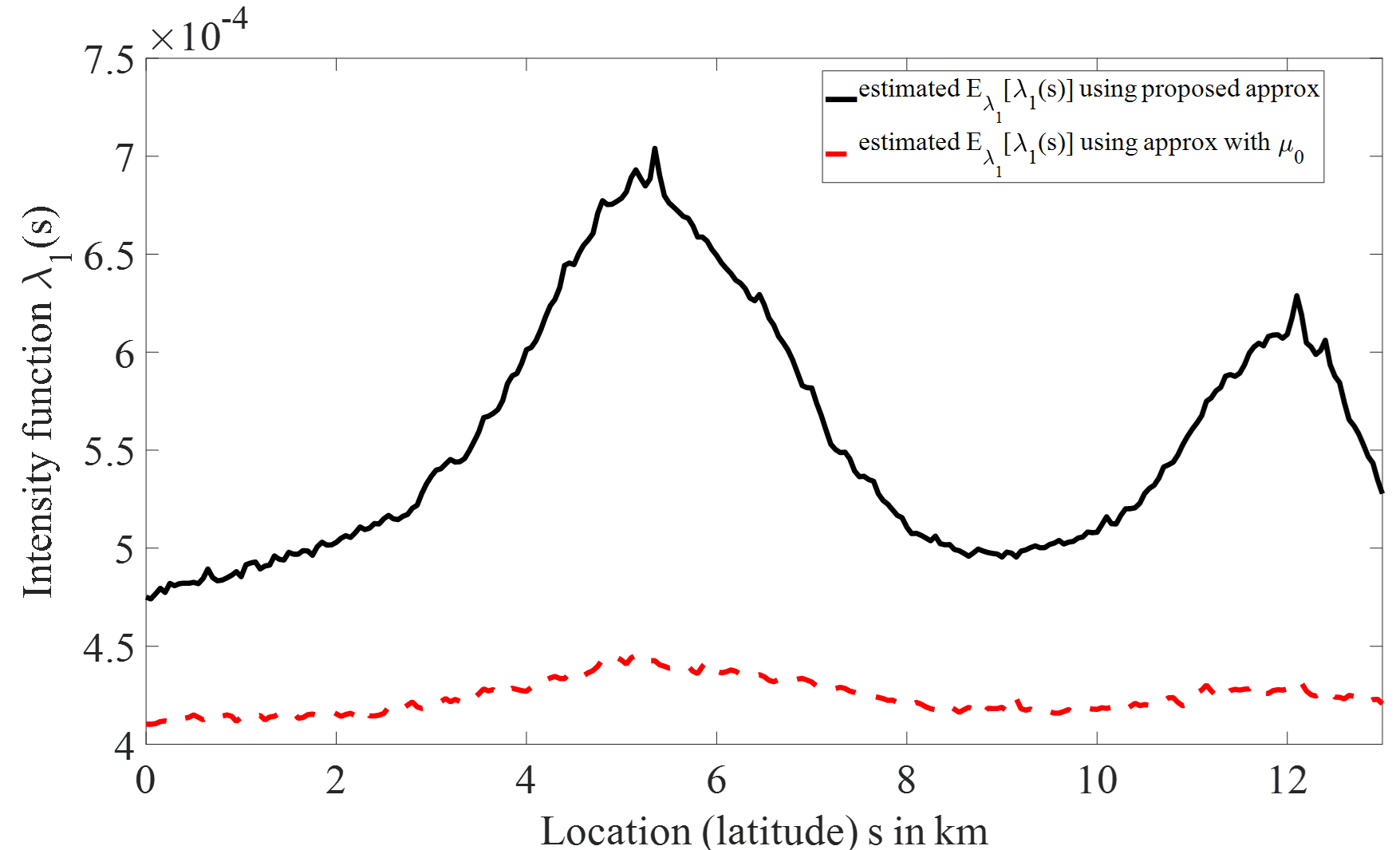}
\caption{Estimated mean intensity functions of potential commission outlier target arrivals, classified using Algorithm~\ref{algo:outlierClassification} based on \eqref{eq: mean approx probability of outlier target arrival} (mean-only approximation) and \eqref{eq: our approx probability of outlier target arrival} (proposed second-order approximation).}
\label{fig:5}
\end{figure}

For implementation of real-time sensor placement, we discretize the 13~km line segment into possible sensor locations at 50-meter intervals, forming a finite set of possible deployment points. We adopt the greedy sensor placement framework from \cite{kim2025variance}, which enables efficient selection of sensor locations in real-time. For sensor modeling, we use the same model as described in Section IV-B of \cite{kim2025variance}.  
We define the probability that a sensor located at position $a_i$ detects an outlier at location $s$ as

\begin{equation}
    \gamma(s, a_i) = \rho \exp\left(-\frac{(a_i - s)^2}{\sigma_l}\right)
    \label{eq:probabilityOfDetection}
\end{equation}

\noindent In this expression, $\rho$ (where $0 \leq \rho \leq 1$) denotes the maximum detection probability, and $\sigma_l$ is a length-scale parameter that controls how quickly the detection capability decreases with spatial distance. For our numerical experiments, we use $\rho = 0.98$ and $\sigma_l = 0.05$, ensuring that the detection probability rapidly declines as the distance between the sensor and the target increases. This configuration enables each sensor to effectively detect targets within an approximate range of 0.56 km.


\begin{figure}
\centering
\includegraphics[scale=0.28]{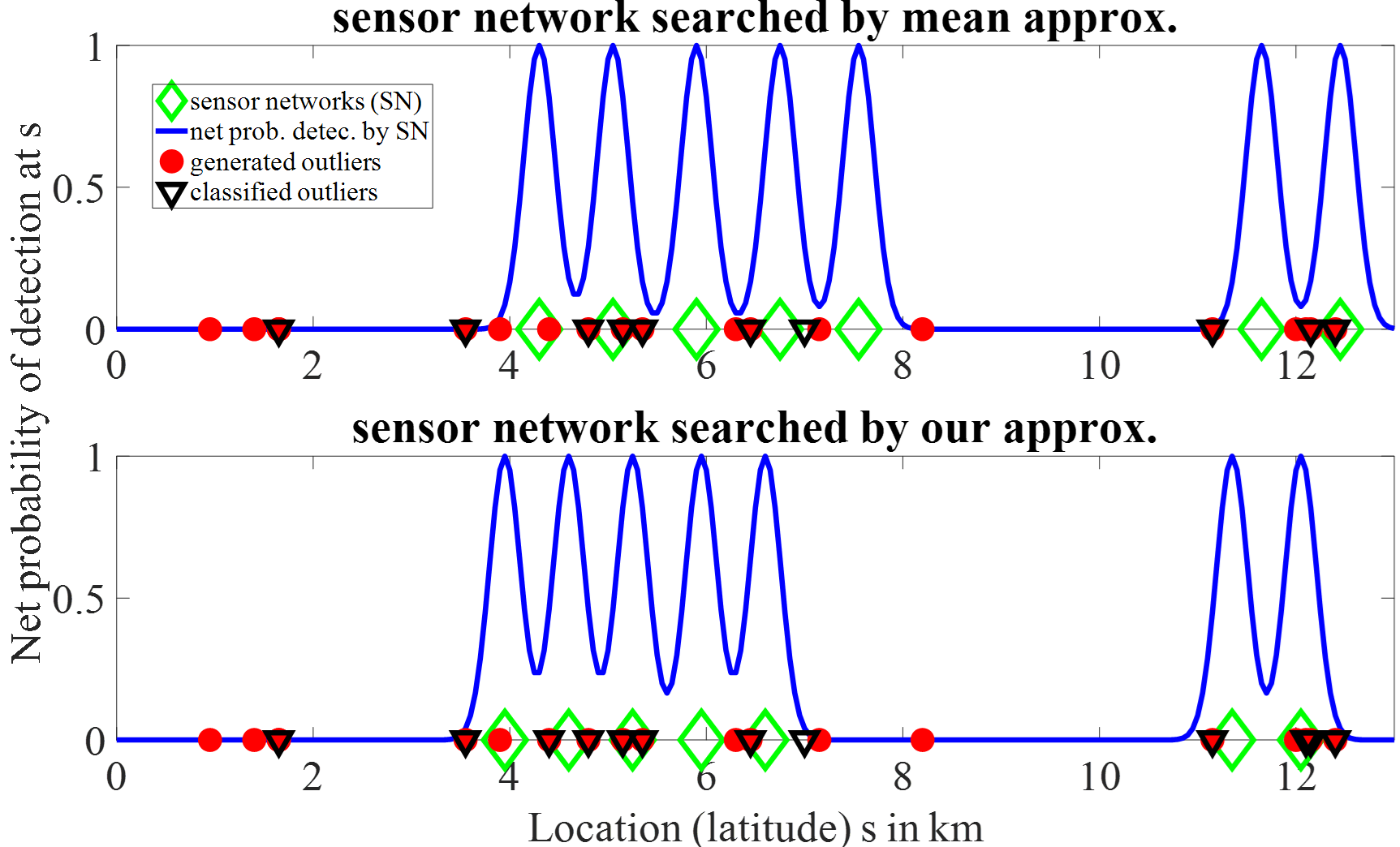}
\caption{Near-optimal placement of 7 sensors for detection of potential commission outlier target arrivals where outliers are classified using (top) \eqref{eq: mean approx probability of outlier target arrival} and (bottom) \eqref{eq: our approx probability of outlier target arrival}. }
\label{fig:6}
\end{figure}

Fig.~\ref{fig:5} shows the estimated mean intensity functions of outlier arrivals, constructed from the classified potential outliers (indicated by upside-down black triangles from Fig.~\ref{fig:2}). Based on these intensity estimates, we select sensor locations that maximize the void probability for detecting potential outliers, as shown in Fig.~\ref{fig:6} with green diamonds.

To evaluate performance, we assess whether the sensor networks can successfully detect the originally generated outliers. Specifically, we compare two configurations: one based on our proposed method and the other using the mean-only approximation. Each network consists of 7 sensors. Our results show that the network designed using the classified data from our proposed method achieves an average improvement of $30.05\%, 15.05\%, 4.34\%$, respectively, for 1-day, 1-week, 1-month long data in outlier detection performance.


The evaluation is performed using a Monte Carlo approach. For each outlier location, we generate random numbers uniformly in the interval $[0, 1]$ and compare them to the probability of detection by the sensor network. The detection probability is computed by subtracting the probability of failing detection by the sensor network, as defined in \eqref{eq:prob. failing detection}, from 1 as shown in Fig.~6 (blue curves). An outlier is considered detected if the random number is less than the corresponding detection probability. This process is repeated 10,000 times to estimate the average outlier detection rate.

\section{Conclusion}

This paper presents a framework for classifying and detecting spatial commission outliers in maritime environments using log-Gaussian Cox processes and seabed acoustic sensor networks. By incorporating both the mean and variance of the normal intensity function, our proposed approximation improves outlier classification accuracy over mean-only approaches. We analytically show that this approximation provides a tighter bound to the true outlier probability. Additionally, we integrate a real-time sensor placement strategy to enhance the detection of classified outliers. The effectiveness of our method is demonstrated using real ship traffic data near Norfolk, Virginia.

While the framework performs well, it assumes full observability of all target arrivals and does not account for cumulative classification errors, which may degrade the accuracy of the estimated outlier intensity over time. As future work, we plan to incorporate cumulative classification uncertainty and develop sensor models capable of jointly handling detection and classification under uncertainty.

\section*{Appendix}
\subsection*{Proof of Lemma 1: Smaller Approximation Error with the Proposed Method}

We aim to show that the proposed second-order approximation in \eqref{eq: our approx probability of outlier target arrival} is closer to the true expected value than the mean-only approximation in \eqref{eq: mean approx probability of outlier target arrival}. That is, we demonstrate that the Jensen gap for our method is smaller in magnitude than that of the mean-only case.

To prove this, it suffices to verify that the right-hand side of the inequality in \eqref{eq: Jensens's inequality 1} is greater than the absolute value of both bounds given in \eqref{eq: Jensens's inequality 2}, under the condition that $\mu_0, \sigma_0, \lambda_1 > 0$.

The right-hand side of \eqref{eq: Jensens's inequality 2} simplifies as
\begin{align*}
    \frac{\sigma_0^2}{(\mu_0 + \lambda_1)^2} - \frac{\lambda_1 \sigma_0^2}{(\mu_0 + \lambda_1)^3}
    = \frac{\mu_0 \sigma_0^2}{(\mu_0 + \lambda_1)^3} > 0,
\end{align*}
which is clearly less than the upper bound of \eqref{eq: Jensens's inequality 1}
\[
    \frac{\sigma_0^2}{(\mu_0 + \lambda_1)^2}.
\]

Next, we compare the magnitude of the lower bound in \eqref{eq: Jensens's inequality 2}
\begin{align*}
    \left| -\frac{\lambda_1 \sigma_0^2}{(\mu_0 + \lambda_1)^3} \right| 
    \leq \left| \frac{\sigma_0^2}{(\mu_0 + \lambda_1)^2} \right|.
\end{align*}

This inequality holds because
\begin{align*}
    \frac{\sigma_0^2}{(\mu_0 + \lambda_1)^2} - \frac{\lambda_1 \sigma_0^2}{(\mu_0 + \lambda_1)^3}
    = \frac{\mu_0 \sigma_0^2}{(\mu_0 + \lambda_1)^3} > 0.
\end{align*}

Hence, the magnitude of the approximation error (Jensen gap) using the proposed method is strictly smaller than that of the mean-only approximation. This confirms that incorporating the variance term yields a closer approximation to the true probability of outlier target arrival at location $s$. \qedsymbol

\bibliographystyle{IEEEtran}
\bibliography{IEEEabrv, main.bib}

\end{document}